\pdfoutput=1

\documentclass[11pt]{article}

\usepackage{acl}

\usepackage{times}
\usepackage{latexsym}

\usepackage[T1]{fontenc}

\usepackage[utf8]{inputenc}

\usepackage{microtype}

\usepackage{inconsolata}

\usepackage{algorithm}
\usepackage{algorithmic}
\usepackage{newfloat}
\usepackage{listings}
\usepackage{inconsolata}
\usepackage{enumitem}
\usepackage{bbm}
\usepackage{algorithm}
\usepackage{algorithmic}
\usepackage{multirow}
\usepackage{multicol}
\usepackage{graphicx}
\usepackage{xcolor}
\usepackage{color}
\usepackage{colortbl}
\usepackage{float}
\usepackage{subcaption}
\usepackage{amsmath}
\usepackage{appendix}

\title{Few-shot Knowledge Graph Relational Reasoning \\ via Subgraph Adaptation}

\author{Haochen Liu \\
  University of Virginia\\
  \texttt{sat2pv@virginia.edu}\vspace{0.15in} \\
  {\bf  Chen Chen} \\
  University of Virginia\\
  \texttt{zrh6du@virginia.edu }\vspace{0.15in} \\
  \And
 {\bf  Song Wang}\\
  University of Virginia\\
  \texttt{sw3wv@virginia.edu}\vspace{0.15in} \\
    {\bf  Jundong Li}\\
  University of Virginia\\
   \texttt{jundong@virginia.edu} 
  }

\begin{document}
\maketitle

\begin{abstract}
Few-shot Knowledge Graph (KG) Relational Reasoning aims to predict unseen triplets (i.e., query triplets) for rare relations in KGs, given only several triplets of these relations as references (i.e., support triplets). This task has gained significant traction due to the widespread use of knowledge graphs in various natural language processing applications. Previous approaches have utilized meta-training methods and manually constructed meta-relation sets to tackle this task. Recent efforts have focused on edge-mask-based methods, which exploit the structure of the contextualized graphs of target triplets (i.e., a subgraph containing relevant triplets in the KG). However, existing edge-mask-based methods have limitations in extracting insufficient information from KG and are highly influenced by spurious information in KG. 
To overcome these challenges, we propose SAFER (\textbf{\underline{S}}ubgraph \textbf{\underline{A}}daptation for \textbf{\underline{FE}}w-shot Relational \textbf{\underline{R}}easoning), a novel approach that effectively adapts the information in contextualized graphs to various subgraphs generated from support and query triplets to perform the prediction.  
Specifically, SAFER enables the extraction of more comprehensive information from support triplets while minimizing the impact of spurious information when predicting query triplets. Experimental results on three prevalent datasets demonstrate the superiority of our proposed framework SAFER.\footnote{Our code is available at \url{https://github.com/HaochenLiu2000/SAFER}.}
\end{abstract}

\section{Introduction}\label{introduction}
\begin{figure}[!t]
  \centering
  \includegraphics[width=1\linewidth]{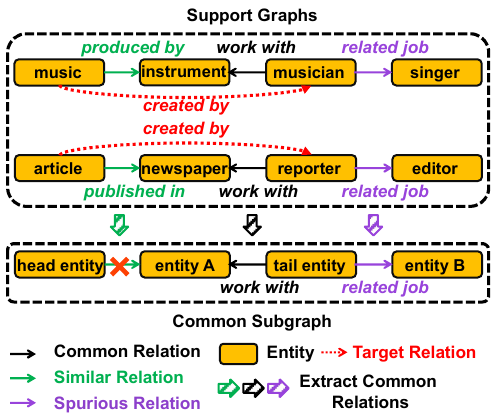}
  \caption{We provide an instance for the two limitations of edge-mask-based methods. In this example, there are two support triplets (\texttt{music, created\_by, musican}) and (\texttt{news article, created\_by, reporter}). When extracting support information by finding the common subgraph, the extraction of edges with similar meanings but in different graphs will fail, and some spurious information will be extracted, which cannot correctly represent the logical pattern of the relation \texttt{created\_by}. }
  \label{fig:challenge1}
  \vspace{-.15in}
\end{figure}

Knowledge Graphs (KGs) consist of many triplets, i.e., (\texttt{head, relation, tail}), which represent specific relationships between real-world entities~\cite{survey2, survey}. These triplets form directed graphs that store knowledge information and can be applied to various knowledge-based tasks~\cite{downstream,wang2023knowledge} such as question answering~\cite{qa1,qa2}, information extraction~\cite{infoextraction2,infoextraction}, program analysis~\cite{programanalysis}, and language model enhancement~\cite{lm1,lm2,lm3}. However, KGs generally cannot encompass all the necessary knowledge triplets required by downstream tasks, as most KGs are severely incomplete~\cite{gmatching}. Therefore, it becomes crucial to complete KGs by inferring potential missing relations between entities. 
In particular, existing works for KG completion~\cite{transe,linkpred,tradcompletion2} often assume the availability of sufficient instances (i.e., triplets) for each relation to be predicted. However, in real-world scenarios, it is common to encounter \emph{few-shot relations}, where only limited instances of triplets with these relations, called \emph{support} triplets, are available.
KGs 
are constantly being updated, for example, by including knowledge from social networks. This often results in new relations with a relatively scarce number of discovered triplets, as the labeling process can be laborious. These new relations are generally known as \textit{few-shot relations}. 
Consequently, predicting new relations with only limited triplets becomes a significant task~\cite{fsgc}. Therefore, it is crucial to perform the \textit{Few-shot KG Relational Reasoning} (Few-shot KGR) task~\cite{gmatching}, which aims to predict the existence of (unseen) \textit{query triplets} of a relation, given a background KG and a set of a limited number of \emph{support triplets} of the relation as the \emph{support set}. 

Currently, there exist two types of approaches for solving the Few-shot KGR task. The first type is \emph{meta-learning-based} methods~\cite{meta-r,fsrl,att}, which utilize the meta-learning framework~\cite{finn2017model} to transfer 
useful knowledge to new KGR tasks~\cite{metalearning} with a limited number of support triplets, to tackle the issue of data scarcity in the target few-shot tasks. 
Nevertheless, 
the distribution of the manually selected target relations plays an important role in these methods, which will result in suboptimal performance if the meta-training sets are not well-designed.
To address this limitation, more recent studies have explored \emph{edge-mask-based} approaches~\cite{csr,sarf}, 
providing an alternative solution to Few-shot KGR tasks. Edge-mask-based methods analyze each support (or query) triplet by first retrieving its contextualized graph, i.e., the subgraph that consists of the head and tail entities of a triplet, and the most relevant entities and relations of the triplet. The subgraph is referred to as the support (or query) graph. 
Then they extract common subgraphs across support graphs in the form of masks that identify edges with shared meanings for predictions on query triplets.

Despite the effectiveness of these works, we argue that there are still two major limitations of edge-mask-based methods.
(1) Existing edge-mask-based approaches 
assume that the largest common subgraph (masks) shared across all support graphs is sufficient to represent the unseen target relation. However, this assumption is difficult to satisfy in certain cases, e.g., when dealing with triplets that involve different but similar relations across other support graphs. As shown in Figure~\ref{fig:challenge1}, on the support graphs of the target relation \texttt{created\_by}, the relations \texttt{produced\_by} and \texttt{published\_in} preserve similar meanings. However, the strategy of learning edge masks fails to harness the valuable insights from these different yet similar relations, resulting in the insufficient extraction of information from \texttt{created\_by}. 
(2) The extracted common subgraph (masks) often contains unrelated spurious information that can negatively impact prediction performance. For example, during the extraction process in Figure~\ref{fig:challenge1} regarding the target relation \texttt{created\_by}, the support graphs may include spurious relations like \texttt{related\_job}, as it can be unhelpful or even misleading when predicting query triplets of relation \texttt{created\_by}. 

To overcome the aforementioned challenges, we propose SAFER (\textbf{\underline{S}}ubgraph \textbf{\underline{A}}daptation for \textbf{\underline{FE}}w-shot Relational \textbf{\underline{R}}easoning), a novel subgraph-based approach that effectively utilizes useful information from support graphs while excluding spurious information. 
In SAFER, we first generate the contextualized graphs of support and query triplets with edge weights representing the importance of each relation for performing relational reasoning. Subsequently, we perform Subgraph Adaptation comprising two crucial modules: \emph{Support Adaptation} and \emph{Query Adaptation}, which aim to extract valuable information from support graphs and exclude spurious information, respectively. In our \emph{Support Adaptation} module, we incorporate information from each support graph into others to enable the adaptation to support graphs with different structures to extract and utilize useful information, e.g., similar relations. 
In our \emph{Query Adaptation} module, we adapt the support information to the structure of the query graph so that spurious information among support graphs can be filtered out in a query-adaptive manner. As a result, we can effectively alleviate the adverse impact of spurious information.
In summary, our contributions in this paper are as follows:
\begin{enumerate}[leftmargin=4.5mm, itemsep=0.01em]
    \item We scrutinize the challenges of few-shot knowledge graph relational reasoning (Few-shot KGR) from the perspective of extracting informative common subgraphs. We also discuss the necessity of tackling the challenges.
    \item We develop a novel Few-shot KGR framework consisting of Subgraph Generation and Subgraph Adaptation. Subgraph Adaptation includes (1) a Support Adaptation (SA) module that enables a more comprehensive extraction of information from the support graphs; (2) a Query Adaptation (QA) module that allows for excluding the influence of spurious information in the extracted information.
    \item We conduct experiments on three prevalent real-world KG datasets of different scales. The results further demonstrate the superiority of SAFER over other state-of-the-art approaches.
\end{enumerate}

\section{Related Work}

\subsection{Meta-learning-based Few-shot KGR}

Meta-learning~\cite{finn2017model,metalearning} is an effective learning paradigm that transfers generalizable knowledge learned from training tasks to new test tasks. Meta-learning necessitates a meta-training set that comprises multiple Few-shot KGR tasks for training purposes and then generalizes learned knowledge to tasks in the meta-test set. 
For example, GMatching~\cite{gmatching} and FSRL~\cite{fsrl}, acquire a universal metric to match query triplets with support triplets~\cite{wang2021reform}. 
The performance of meta-learning is significantly influenced by the quality of the manually created meta-training set. Moreover, the meta-training set is sampled from the same distribution as the meta-test set, which is impractical in practice~\cite{csr}. To overcome these problems, some alternative studies based on subgraph structures are proposed to tackle the Few-shot KGR task.

\subsection{Edge-mask-based Few-shot KGR}

Edge-mask-based methods, such as CSR~\cite{csr} and SARF~\cite{sarf}, consider the few-shot relational reasoning task as an inductive reasoning problem~\cite{inductive,subgraph}, which relies on the relevant relations(i.e., edges) of the triplet
~\cite{rule1,path1,rule2} in KG to perform the prediction. These methods employ an encoder-decoder model to encode the shared subgraphs of support samples (masks), i.e., common subgraphs in KG that connect the two entities of the triplets, into an embedding representing the target relation. The decoder uses the embedding to reconstruct the edge masks in a query graph showing the shared edges. 
These approaches take advantage of the edge structure to perform reasoning. However, these methods have the limitation that the largest common subgraph among support graphs may lose some of the relation's logical patterns, and the spurious information extracted will detrimentally affect the prediction. In this paper, our approach uses a novel adaptation process to address the shortcomings of incomplete utilization of structure information in edge-mask-based methods.

\section{Problem Formulation}

We study the problem of \emph{Few-shot Knowledge Graph Relational Reasoning}, i.e., Few-shot KGR~\cite{gmatching,meta-r}. We first denote the background KG 
as $\mathcal{G}=(\mathcal{E}, \mathcal{R}, \mathcal{T})$, where $\mathcal{E}$ and $\mathcal{R}$ are sets of entities and relations. $\mathcal{T}=\{(h,r,t)|h,t\in\mathcal{E},r\in\mathcal{R}\} $ represents the facts as triplets, each of which contains a head entity, a tail entity, and a relation. For a new target relation $r'\notin\mathcal{R}$, we are given a support set $S_{r'}$ with $K$ triplets $\{(h_i,r',t_i)\}_{i=1}^K$ of $r'$, where $h_i,t_i\in \mathcal{E}$. The number of triplets in the support set $K$ is relatively small ($K\leq5$). With $S_{r'}$ as the reference, we aim to predict tail entities, given a head entity $h_q$, i.e., $(h_q,r',?)$. 
 There are usually multiple candidates of the tail entity that need to be scored and ranked. Then the candidate with the highest score is considered as the prediction result. 
So we will consider the query triplet $(h_q,r',c)$ ($c$ is a candidate) as a full triplet to score.

\section{Methodology}

\begin{figure*}[!t]
  \centering  \includegraphics[width=1\linewidth]{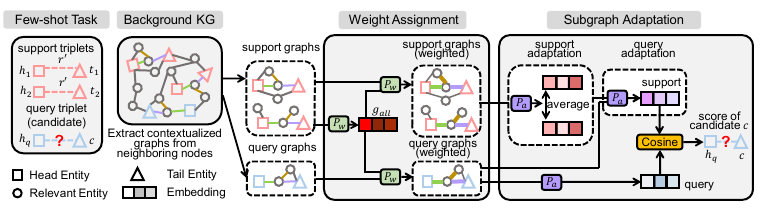}
  \caption{The framework of SAFER, which shows the scoring pipeline for a query tail candidate $c$ of target relation $r'$. 
  We represent the same relations in colors, while the gray relations are all different. 
  We first extract the contextualized graph of each support and query triplet and assign weights to all edges using an aggregation process $P_w$ (the width of edges represents weights). Then we apply another aggregation process $P_a$ and two adaptation operations to perform support information extraction and query candidate scoring.}
  \label{fig:framework}
  \vspace{-0.2in}
\end{figure*}

In this section, we introduce details of our proposed framework SAFER. 
As illustrated in Figure~\ref{fig:framework}, for each support (or query) triplet, we first extract a support (or query) graph from the background KG and assign weights for each edge on the graph. Then we conduct Subgraph Adaptation on the generated support and query graphs and finally achieve the prediction score for a query triplet.

\subsection{Retrieving Contextualized Graphs}
To obtain structural information for the unseen target relation, we utilize the contextualized graphs of support and query triplets, i.e., \emph{support graphs} and \emph{query graphs}. 
Contextualized graphs are generated based on the enclosing subgraph strategy proposed by~\cite{subgraph2,subgraph}. We introduce how to construct contextualized graphs in Appendix~\ref{contextualized}. 

\subsection{Edge Weight Assignment}\label{weight}

After acquiring the contextualized graph, we propose to assign weights to all edges on the contextualized graphs based on their importance to the target relation. We assign the weight $w_e$ for each edge $e$ by incorporating information from all support graphs to determine the importance, such that we can effectively leverage the information within all relations.


Specifically, 
we leverage the PathCon~\cite{messagepassing} model to extract structural information and calculate the edge weights, as it can measure graph isomorphism. While edge-mask-based methods apply the model repeatedly between any two graphs to get the masks, we only apply it to get an overall embedding $g_{all}$ of all support graphs.

We define an aggregation process $P_w$ with $L$ iterations as follows:
\begin{equation}\label{equation2}
b_v^i=\frac{1}{1+|\{e|e\in N(v)\}|}\sum_{e\in N(v)} b_e^i,
\end{equation}
\begin{equation}\label{equation3}
r_v^i=b_v^i\|\mathbbm{1}(v=h)\|\mathbbm{1}(v=t),
\end{equation}
\begin{equation}\label{equation4}
    b_e^{i+1}=f(r_u^i\|r_v^i\|b_e^i),
u,v\in N(e),
\end{equation}
where $b_e^i$ (or $b_v^i$) is the learned edge (or node) embedding in iteration $i$. $N(v)$ is the set of all neighboring edges of $v$. 
$f$ is a neural network (NN) consisting of both non-linear and linear layers. $\|$ denotes the concatenation of two vectors (or scalars).
In particular, Eq.~(\ref{equation2}) aggregates the embeddings of neighboring edges of each node. Then Eq.~(\ref{equation3}) adds the label of head and tail so that the information of a node's relative position to head and tail can be considered. Eq.~(\ref{equation4}) updates all edge embeddings based on the current embedding of the edge and its two end nodes. 

In the first step, 
we initilize $b_e^0$ with the pretrained relation embedding $v_e$ of the relation on edge $e$. We define the embedding of $G$ as follows:
\begin{equation}\label{equation5}
g(G)=\text{MaxPool}(b_v^L)\|b_h^L\|b_t^L,
\end{equation}
where $\text{MaxPool}(b_v^L)$ is the max-pooling of all node embeddings in $G$.

In the second step, similarly, we apply $P_w$ again to acquire the weights of edges in both the support graphs and the query graphs. Additionally, we use the average of the embeddings of all support graphs $g_{all}$ from the first step as an input to incorporate the overall information in the support set and initialize $b_e^0$ as $v_e\|g_{all}$. Here $g_{all}$ is defined as follows:
\begin{equation}
    g_{all}=\frac{1}{K}\sum_kg(G_s^k).
\end{equation}
Here $G_s^k$ is the $k$-th support graph. We use another $f$ in this step. 
Then we perform $P_w$ on the target graph $G$. Finally, we calculate the weight $w_e$ of edge $e$:
\begin{equation}\label{equation7}
    w_e=\frac{1}{1+\exp(-\text{Linear}(b_e^L))},
\end{equation}
where $\text{Linear}(\cdot)$ is a linear layer, and $w_e$ will serve as the edge weight of $e$ in the subsequential adaptation modules.

Note that weight assignment does not rely on specific loss functions or ground-truth definitions for edge weights. Instead, it is trained in an end-to-end manner along with other modules in the subsequent sections. All edges in the support graphs can contribute to the subsequential adaptation modules based on the weight. 

\subsection{Subgraph Adaptation}

In this subsection, we introduce the process of our Subgraph Adaptation module, including \emph{Support Adaptation} (SA) and \emph{Query Adaptation} (QA). 

After obtaining the edge-weighted support graphs and query graphs, we 
achieve embeddings that contain the information from different subgraphs by aggregations. 
While performing the aggregations, we further adapt graph information to all support and query graphs to perform SA and QA. We first define an $L$-iteration aggregation process $P_a$, which is utilized in both SA and QA:
\begin{equation}\label{equation8}
a_v^i(k)=\frac{1}{1+\sum_{e\in N(v)} w_e(k)}\sum_{e\in N(v)} b_e^i(k)\cdot w_e(k),
\end{equation}
\begin{equation}\label{equation9}
\renewcommand{\arraystretch}{1.5}
\setlength{\arraycolsep}{2.2pt}
\begin{aligned}
        b_v^i(k)= 
        \left\{\begin{array}{ll}T_{SA}(\{a^i_v(m)\}_{m=1}^{K}),
        &
        \text{if}\ \text{} \text{SA}, \\
        T_{QA}(a^{i}_v(k), \{b^i_t(m)\}_{m=1}^K;\lambda), 
        &\text{if}\ \text{} \text{QA},\end{array} \right.
\end{aligned}
\end{equation}
\begin{equation}
r_v^i(k)=b_v^i(k)\|\mathbbm{1}(v=h)\|\mathbbm{1}(v=t),
\end{equation}
\begin{equation}
    b_e^{i+1}(k)=f(r_u^i(k)\|r_v^i(k)\|b_e^i(k)),
u,v\in N(e),
\end{equation}
where $k$ indicates that a term is calculated on the $k$-th support graph, and it can be replaced by $q$ to represent the value on a query graph in \emph{Query Adaptation} (e.g., $a^i_v(q)$ and $b^i_v(q)$). $N(v)$ is the set of all neighboring edges of node $v$. $w_e$ is the weight of edge $e$. $a_v^i$ is the aggregation output of node $v$ at iteration $i$. Here Eq.~(\ref{equation8}) aggregates the embeddings of all neighboring edges of each node based on edge weights. $b_v^i$ (or $b_e^i$) is the learned node (or edge) embedding in iteration $i$. The adaptation steps are $T_{SA}(\cdot)$ (for SA) and $T_{QA}(\cdot)$ (for QA), and the details will be introduced in the following subsections. $f$ is a neural network (NN) consisting of non-linear and linear layers acting in both SA and QA. $\lambda$ is a hyperparameter used in QA to be introduced.
Note that we initialize $b_e^0(k)$ with the pretrained embedding of the relation on edge $e$ to incorporate more information.

\subsubsection{Support Adaptation}
To extract valuable information from all support graphs and reduce the omissions of information, we propose the \emph{Support Adaptation} (SA) strategy that enables the incorporation of information from all support graphs when learning the embedding for each support graph. During aggregation on each graph, we average the learned embeddings of the tail entities in all support graphs after each iteration to absorb beneficial information from all other support graphs. In particular, we choose to average the embeddings of \textit{tail} entities (instead of other entities), because the tail entity preserves the most crucial information for the prediction of the target relation. The averaged embedding will be used to update embeddings of all edges connected to tail entities in all support graphs. This strategy ensures the transfer of relational information from one support graph to various others, thereby enabling adaptation to structures of different support graphs during subsequent aggregation steps.
In this way, all edges in the support graph can contribute to SA based on their weights.

In SA, we apply $P_a$  to all $K$ support graphs for $L$ iterations. 
$T_{SA}(\cdot)$ is defined as
\begin{equation}\label{equation10}
\renewcommand{\arraystretch}{1.2}
\setlength{\arraycolsep}{12.5pt}
\begin{aligned}
        &T_{SA}(\{a^i_v(m)\}_{m=1}^{K}) =\\ &\left\{
\begin{array}{ll}
\frac{1}{K}\sum_{m=1}^Ka^i_t(m), & \text{if}\  v=t,\\[6pt]
a_v^i(k), & \text{otherwise}. 
    \end{array}\right.
\end{aligned}
\end{equation} 

Via Eq.~(\ref{equation10}), we manage to incorporate information from other support graphs when performing aggregation on each support graph. Generally, if the information from a specific relation in a support graph can be easily propagated on another support graph with a different relation, we can infer that these two relations maintain similar meanings. Therefore, our 
SA strategy allows for extracting relevant relations (e.g., different yet similar relations) among support graphs.



\subsubsection{Query Adaptation}

\emph{Query Adaptation} (QA) is the subsequent module that can exclude the influence of spurious information extracted by the 
SA module. Generally, we predict the score of a query triplet by comparing the similarity between information learned from the query graph and the support graphs.
To deal with the presence of spurious information across query and support graphs, 
our QA module adapts the tail node embeddings in support graphs to the structure of the query graph. In this manner, the support information unhelpful for query scoring will be filtered out,  
due to different structures between support graphs and query graphs. Then we calculate the score of a query triplet by comparing the filtered support embedding with the embedding of the query graph.

To perform QA, we apply the aggregation process $P_a$ to the query graph of the query triplet candidate. $T_{QA}(\cdot)$ is defined as follows:
\begin{equation}\label{equation12}
\renewcommand{\arraystretch}{1.2}
\setlength{\arraycolsep}{2.5pt}
\begin{aligned}
      &T_{QA}(a_v^i(q),\{b^i_t(m)\}_{m=1}^K;\lambda)=\\
      &\left\{
\begin{array}{ll}
 (1-\lambda)\cdot a_t^i(q)+\frac{\lambda}{K}\sum_{m=1}^Kb^i_t(m),&\text{if}\ v=t,\\[6pt]
    a_v^i(q), &\text{otherwise}. 
   \end{array}\right. 
\end{aligned}
\end{equation}
Here $\lambda\geq0$ is a hyperparameter of QA, which shows the ratio of incorporation of extracted support information and the information from the query graph. In this manner, we perform aggregation for support information on the query graph. As a result, our 
QA module can exclude the influence of spurious information in support graphs, thus achieving more precise prediction results.

To perform prediction for a query triplet, we compare two embeddings, $E_s$ and $E_q$, which involve (filtered) support information and query information, respectively.
Specifically, we define 
\begin{equation}
E_s=T_{QA}(a_t^L(q),\{b^L_t(m)\}_{m=1}^K;\lambda)
\end{equation}
as the result of the filtered support information with $\lambda>0$ obtained from Eq.~(\ref{equation12}). 
For $E_q$, we perform $P_a$ with $\lambda=0$ to ensure that there is no incorporation of support information. We define $E_q$ as follows:\begin{equation}
E_q=T_{QA}(a_t^L(q),\{b^L_t(m)\}_{m=1}^K;0).
\end{equation} As the calculation of $E_q$ does not involve information from support graphs, $E_q$ only contains the query information. 
Additionally, we concatenate the average of pretrained embeddings of all support and query tail entities to $E_s$ and $E_q$, respectively, so that the pretrained entity embedding can also contribute to the scoring.
In particular, we use the cosine similarity between $E_s$ and $E_q$ to measure the score of a query candidate, denoted as 
\begin{equation}
    s(t_q)=\text{cos}(E_s\|\frac{1}{K}\sum_{k=1}^K v_{t_{s,k}},E_q\|v_{t_q}),
\end{equation}
where $s(t_q)$ is the score for $t_q$, i.e., the tail entity of the query triplet. $t_{s,k}$ is the tail entity of the $k$-th support triplet. We use $v_{t_{s,k}}$ (or $v_{t_q}$) to denote the pretrained node embedding of $t_{s,k}$ (or $t_q$).
Note that both $E_s$ and $E_q$ are solely acquired via aggregation on the query graph. This ensures 
exclusion of spurious information in support graphs, thus achieving more precise scoring results.

\subsection{Training Objective}

To train the overall SAFER framework, we leverage contrastive learning with positive samples (i.e., same relation in support and query triplets) and negative samples (i.e., different relations in support and query triplets). Specifically, we use the Margin Ranking Loss:
\begin{equation}
    \mathcal{L}=\max(s_{neg}-s_{pos}+\gamma,0),
\end{equation}
where $s_{pos}$ and $s_{neg}$ are scores of the positive sample and the negative sample, respectively. $\gamma\in\mathbbm{R}$ is a hyperparameter utilized to control the margin that separates positive and negative samples.

\section{Experiments}

In this section, we elaborate on the experiments for evaluating our proposed framework. 

\subsection{Experimental Settings}
\subsubsection{Datasets} We evaluate our framework and other baselines on three real-world Few-shot KGR datasets, generated based on NELL~\cite{nell}, FB15K-237~\cite{fb15k}, and ConceptNet~\cite{conceptnet}, respectively. The NELL dataset is a subset of NELL-One~\cite{meta-r} by selecting the relations that have between 50 and 500 triples as few-shot tasks. For FB15K-237 and ConceptNet, we select the fewest 30 and 2 appearing relations as test few-shot tasks, respectively, following~\cite{dataset} and~\cite{meta-r}. Table~\ref{dataset} lists the statistics of all three datasets. 

 

\subsubsection{Evaluation Metrics} We perform the evaluation for our framework and all baselines by calculating the scores for query candidates of each test instance using the standard ranking metrics. In particular, we utilize the Mean Reciprocal Ranking (MRR) and Hits@h. The MRR measures the average reciprocal rank of the correct candidate in the ranking of all candidates, where a higher value indicates better performance. We also compute the Hits@h value, which measures the percentage of the correct candidates ranked within the top $h=\{1,5,10\}$ positions. 
In evaluation, each correct candidate in the test set is paired with 50 other candidate negative triplets. 

\subsubsection{Baselines} We compare our framework with existing Few-shot KGR methods, including MetaR~\cite{meta-r}, FSRL~\cite{fsrl}, CSR-OPT~\cite{csr}, CSR-GNN~\cite{csr}, SARF+Learn~\cite{sarf}, and SARF+Summat~\cite{sarf}. For meta-learning-based methods, the training is achieved by randomly sampling tasks from the KG rather than the meta-training split that is originally provided, 
to avoid the influence of manually constructed meta-training sets.

\begin{table}[!t]\centering
\caption{\label{dataset}
Statistics of three Few-shot KGR datasets.
}\vspace{-0.1in}
\renewcommand{\arraystretch}{1.3}
\setlength{\tabcolsep}{4.8pt}
\scalebox{0.8}{
\begin{tabular}{c|cccc}
\hline
Dataset & \# Entities & \# Relations & \# Edges & \# Tasks \\ 
\hline
NELL & 68,544 & 291 & 181,109 & 11 \\ 
FB15K-237 & 14,543 & 200 & 268,039 & 30 \\
ConceptNet & 790,703 & 14 & 2,541,996 & 2 \\ 
\hline
\end{tabular}
}
\vspace{-.15in}
\end{table}

\begin{table}[!t]\centering
\caption{\label{result}
Performance comparison of different KG datasets. The best and second-best results are shown in \textbf{bold} and \underline{underlined}, respectively.
}
		\renewcommand{\arraystretch}{1.3}\setlength\tabcolsep{3.72pt}
\scalebox{0.73}{
\begin{tabular}{c|c|cccc}\hline
Dataset& Method& MRR & Hits@1 & Hits@5 & Hits@10 \\\hline 
\multirow{7}{*}{NELL}&
MetaR & 0.471 & 0.322 & 0.647 & 0.763 \\
  &FSRL & 0.490 & 0.327 & 0.695 & 0.853 \\
  &CSR-OPT & 0.463 & 0.321 & 0.629 & 0.760 \\
  &CSR-GNN & 0.577 & 0.442 & 0.746 & 0.858 \\
  &SARF+Learn & \underline{0.627} & \underline{0.493} & \underline{0.798} & \underline{0.877}\\
  &SARF+Summat & 0.626 & 0.493 & 0.797 & 0.875 \\
  &\cellcolor{gray!21}SAFER (ours) &\cellcolor{gray!21}\textbf{0.674} &\cellcolor{gray!21}\textbf{0.560} &\cellcolor{gray!21}\textbf{0.812} &\cellcolor{gray!21}\textbf{0.887} \\\hline 

\multirow{7}{*}{FB15K-237}&
MetaR & \textbf{0.805} & \textbf{0.740} & \textbf{0.881} & \textbf{0.937} \\
  &FSRL & 0.684 & 0.573 & 0.817 & 0.912 \\
  &CSR-OPT & 0.619 & 0.512 & 0.747 & 0.824 \\
  &CSR-GNN & 0.781 & 0.718 & 0.851 & 0.907 \\
  &SARF+Learn & 0.779 & 0.718 & 0.846 & 0.905\\
  &SARF+Summat & 0.753 & 0.688 & 0.814 & 0.884 \\
  &\cellcolor{gray!21}SAFER (ours) &\cellcolor{gray!21}\underline{0.793} &\cellcolor{gray!21}\underline{0.728} &\cellcolor{gray!21}\underline{0.860} &\cellcolor{gray!21}\underline{0.914}\\\hline 

\multirow{7}{*}{ConceptNet}&
MetaR  & 0.318 &0.226 & 0.390 & 0.496 \\
  &FSRL  & 0.577 & 0.469 & 0.695 & 0.753\\
  &CSR-OPT & 0.559 & 0.450 & 0.692 & 0.736\\
  &CSR-GNN & 0.606 & 0.496 & \textbf{0.735} & \textbf{0.777}\\
  &SARF+Learn & 0.613 & 0.511 & \underline{0.731} & \underline{0.771}\\
  &SARF+Summat & \underline{0.624} & \underline{0.527} & 0.729 & 0.768\\
  &\cellcolor{gray!21}SAFER (ours) &\cellcolor{gray!21}\textbf{0.638} &\cellcolor{gray!21}\textbf{0.564}\cellcolor{gray!21}&\cellcolor{gray!21}0.721 &\cellcolor{gray!21}0.743\\\hline 
\end{tabular}
}
\vspace{-.1in}
\end{table}

\subsection{Performance Comparison}

 
The detailed settings of our experiments are in Appendix~\ref{setting}. 
We evaluate SAFER along with other methods on the three datasets. For baseline performance, we use the experimental results from~\cite{csr} and~\cite{sarf}. Table~\ref{result} shows that our method outperforms baselines in most cases. In NELL and ConceptNet, the improvement of SAFER on the testing MRR is $7.67\%$ and $2.24\%$. The improvement of Hit@1 is $13.59\%$ and $7.02\%$. On FB15K-237, our method is the second best, while being very close to MetaR. The reason is that FB15K-237 contains a large number of relations whose contextualized graphs contain only one triplet, and thus the methods based on the subgraphs' structure (i.e., CSR, SARF, SAFER) are limited in performance. 

Compared to baselines, SAFER shows more significant advantages in MRR and Hits@1. This is because, for the query candidates with high scores, the information provided by the support and query graphs will be similar. Thus, the spurious information in support graphs will more seriously impact the scoring. Nevertheless, our process avoids spurious information in support graphs, which contributes more to the detailed comparison between high-score samples. Thus, SAFER achieves a more precise scoring result.

		\begin{figure}[!t]
		\centering
\captionsetup[sub]{skip=-1pt}
{\includegraphics[width=0.23\textwidth]{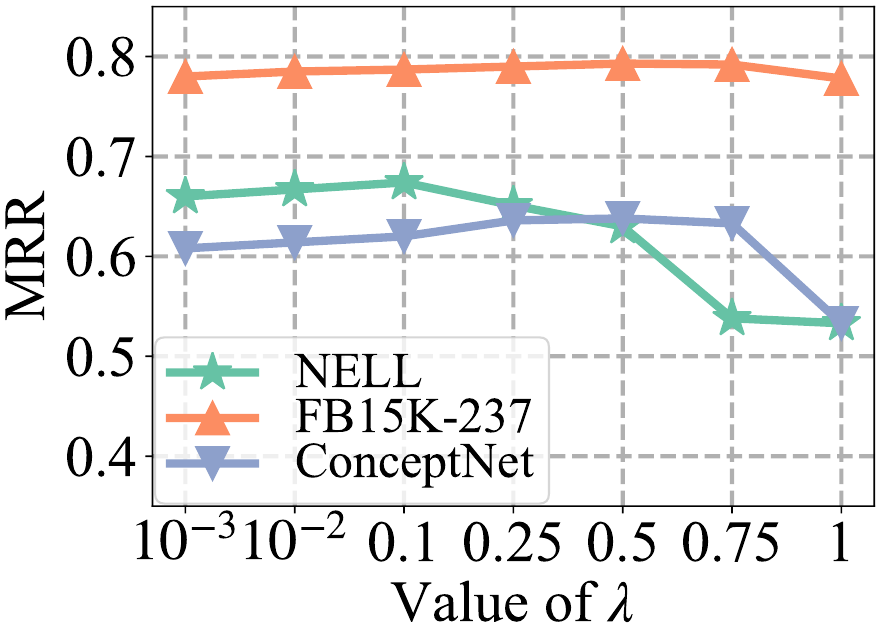}}
{\includegraphics[width=0.23\textwidth]{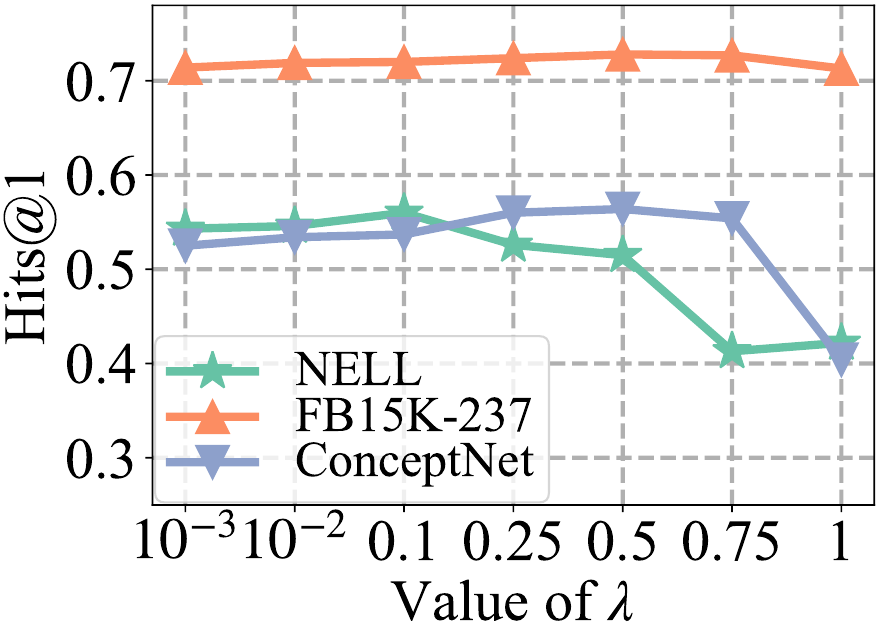}}
		\caption{The performance of our proposed method SAFER with different $\lambda$.  }
  \label{lambda}
  \vspace{-.15in}
	\end{figure}

\subsection{Hyperparameter Study}

The value of $\lambda$ balances the removal of spurious information and the prevention of over-filtering in QA. 
To study the impact of $\lambda$, we conduct experiments with different values of $\lambda$, ranging from 
0.001 to 1. 
The experimental results are presented in Figure~\ref{lambda}. In general, these results indicate that different datasets have different optimal values of $\lambda$. For both MRR and Hits@1, the optimal $\lambda$ is 0.1 for NELL and 0.5 for FB15K-237 and ConceptNet. When $\lambda=1$, the scoring process is actually a direct comparison between the outputs $b_t^L$ of support graphs and the query graph in $P_a$ without any adaptation. 
In this case, the results are much worse than the optimal results, which demonstrates the strength of our QA 
module. For the NELL dataset, the optimal value of $\lambda$ is much smaller because the candidates in NELL have more complex subgraphs and thus require a more precise comparison of the detailed local features.


\subsection{Ablation Study}

\begin{table}[!t]\centering
\caption{\label{ablation}
Ablation study on three datasets. The best results are shown in \textbf{bold}.
}
		\renewcommand{\arraystretch}{1.3}\setlength\tabcolsep{5.2pt}
\scalebox{0.75}{
\begin{tabular}{c|c|cccc}\hline
Dataset& Method& MRR & Hits@1 & Hits@5 & Hits@10 \\\hline 
\multirow{4}{*}{NELL}&\cellcolor{gray!21}SAFER &\cellcolor{gray!21}\textbf{0.674} &\cellcolor{gray!21}\textbf{0.560} &\cellcolor{gray!21}\textbf{0.812} &\cellcolor{gray!21}\textbf{0.887}\\
 &SAFER$\backslash$W & 0.546 & 0.428 & 0.683 & 0.752\\
  &SAFER$\backslash$S & 0.575 & 0.434 & 0.753 & 0.832 \\
  &SAFER$\backslash$Q & 0.533 & 0.422 & 0.659 & 0.715 \\\hline 
\multirow{4}{*}{FB15K-237}&\cellcolor{gray!21}SAFER &\cellcolor{gray!21}\textbf{0.793} &\cellcolor{gray!21}\textbf{0.728} &\cellcolor{gray!21}\textbf{0.860} &\cellcolor{gray!21}\textbf{0.914} \\
 &SAFER$\backslash$W & 0.761 & 0.689 & 0.840 & 0.901 \\
  &SAFER$\backslash$S & 0.761 & 0.688 & 0.841 & 0.901 \\
  &SAFER$\backslash$Q &0.778 & 0.713 & 0.846 & 0.905 \\\hline 
\multirow{4}{*}{ConceptNet}&\cellcolor{gray!21}SAFER &\cellcolor{gray!21}\textbf{0.638} &\cellcolor{gray!21}\textbf{0.564} &\cellcolor{gray!21}\textbf{0.721} &\cellcolor{gray!21}\textbf{0.743}\\
 &SAFER$\backslash$W& 0.474 & 0.331 & 0.632 & 0.729\\
  &SAFER$\backslash$S & 0.510 & 0.399 & 0.629 & 0.728\\
  &SAFER$\backslash$Q &0.533 & 0.404 & 0.710 & 0.742\\\hline 
\end{tabular}
}
\vspace{-.15in}
\end{table}

In this subsection, we conduct an ablation study to evaluate the contributions of the three modules in SAFER: Weight Assignment, Support Adaptation, and Query Adaptation. 
In particular, we remove one module in SAFER each time and report the performance of the revised model on all three datasets. For SAFER$\backslash$W, we directly set the weight $w_e=1$ for all edges to remove the Weight Assignment module. For SAFER$\backslash$S, we remove the SA module 
by removing the averaging in each iteration of $P_{a}$ and only using the average of its final outputs as the support embedding. For SAFER$\backslash$Q, we set $\lambda=1$ to change the scoring into a direct comparison between the outputs $b_t^L$ of support graphs and the query graph in $P_a$ without QA. 

The results of the ablation study, presented in Table~\ref{ablation}, validate the effectiveness of all modules in SAFER. 
Removing the Weight Assignment module significantly decreases the MRR metric. This demonstrates the importance of the weights in the data preparation. 
Furthermore, removing the SA 
module leads to a decrease in all evaluation metrics. This is because, at each iteration of the $P_a$, the aggregations of embeddings from other graphs can emphasize relevant relations in the support graphs. Without this module, the adaptation process becomes a simple average of the final outputs of $P_a$ of all support graphs, resulting in a loss of emphasis on critical information. 
Furthermore, the results highlight the importance of the QA 
module, particularly in terms of MRR and Hit@1 that reflect the similarity between high-score candidates and support samples. By filtering the support information, QA 
ensures that only relevant, and useful information from the support graph is retained. This prevents the inclusion of spurious information within the predefined limits (e.g. common subgraph), thus ultimately contributing to improved performance.

\subsection{Case Study}

\begin{figure}[!t]
  \centering
  \includegraphics[width=1\linewidth]{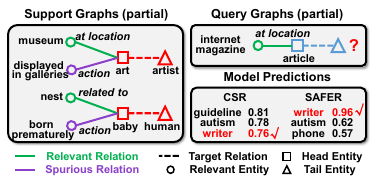}
  \caption{An instance on dataset ConceptNet using the edge-mask-based method CSR and our method SAFER. The figure shows part of support and query graphs and the scores of the 3-top candidates of the two methods. 
  The shown edges prove the limitation of the extraction of common subgraphs in edge-mask-based methods.} 
  \label{case}
  \vspace{-.22in}
\end{figure}

In this section, we study the case that, in existing edge-mask-based methods, the extracted masks (common subgraph) could not correctly represent the target relation all the time. We use a real example in the ConceptNet test set to demonstrate the limitations of extracting common subgraphs to represent the logical pattern of the target relation.

We consider the 2-shot relational reasoning task with two support triplets (\texttt{art}, \texttt{created\_by}, \texttt{artist}) and (\texttt{babies}, \texttt{created\_by}, \texttt{humans}), along with a query triplet (\texttt{article}, \texttt{created\_by}, \texttt{writer}). Here we use an example with both two cases of extracted spurious relations and unextracted relevant relations in the edge-mask-based methods to showcase the two limitations of edge-mask-based methods, as shown in Figure~\ref{case}. 
In the observed support graphs, we can identify two edges of relations \texttt{at\_location} and \texttt{related\_to} as similar but unshared information, and edges of relation \texttt{action} as spurious information. 

Regarding the prediction results, our approach SAFER ranks the true answer of the correct tail entity \texttt{writer} as first of all candidates, whereas the CSR model ranks it as third of all candidates. In the scoring result of CSR, incorrect candidates \texttt{guideline} and \texttt{autism} both receive higher scores than \texttt{writer}. This study shows that our SAFER can actually solve the two limitations of existing edge-mask-based methods in information extraction and processing.


\section{Conclusion}

In this paper, we introduce SAFER, a novel approach designed to address the challenges in Few-shot Knowledge Graph Relational Reasoning (Few-shot KGR). SAFER overcomes the limitations of existing methods by extracting useful information while excluding spurious information. 
We first generate edge-weighted subgraphs of triplets to retrieve useful information from the knowledge graph. With the generated subgraphs, we perform Support Adaptation, which enables the incorporation of useful information that is difficult to extract (e.g., different yet similar relations). Subsequently, our Query Adaptation module filters out spurious information that is easily extracted (e.g., unhelpful relations that are shared across support graphs). Experimental evaluations on three datasets demonstrate the superiority of SAFER over other state-of-the-art baselines under different evaluation metrics. In summary, our work provides valuable insights into the potential of subgraph adaptation to improve performance on Few-shot KGR tasks.

\section{Acknowledgement}
This work is supported in part by the National Science Foundation under grants (IIS-2006844, IIS-2144209, IIS-2223769, CNS2154962, and BCS-2228534), the Commonwealth Cyber Initiative Awards under grants (VV-1Q23-007, HV2Q23-003, and VV-1Q24-011), the JP Morgan Chase Faculty Research Award, and the Cisco Faculty Research Award.


\bibliography{anthology}

\newpage
\mbox{}
\newpage
\appendix
\section{Appendix}
\subsection{Retrieving Contextualized Graphs}\label{contextualized}

In this section, we introduce how we retrieve contextualized graphs from a triplet. 

Contextualized graphs are generated based on the enclosing subgraph strategy proposed by~\cite{subgraph2,subgraph}. Specifically, for a given triplet $(h,r,t)$, we first sample the nodes within $n$-hop undirected neighbors of both the head entity $h$ and the tail entity $t$ from the background KG. To include sufficient nodes for logic extraction, we also perform random sampling from all neighbors of $h$ and $t$.
The resulting contextualized graph is induced by all selected nodes and their connections. It should be noted that the specific value of $n$ is determined based on the density of the KG. In particular, these contextualized graphs can capture the local structure and relevant entities surrounding the support and query triplets, thus allowing us to extract valuable information for the relational reasoning task.

\subsection{Experimental Settings}\label{setting}
In this section, we delve into a more comprehensive exposition of our experimental setups, including detailed parameter settings, as applied to the three distinct real KG datasets.

In our experiments, we have employed 3-shot relational reasoning tasks across all three datasets. For the NELL dataset, we set $n=2$ hops, whereas, for both the FB15K-237 and ConceptNet datasets, we use $n=1$ hop when generating the contextualized graphs of their respective triplets.

Regarding the neural network $f$, we have incorporated three distinct neural networks for the first and second steps of weight assignment and the adaptation module. The overall iteration of all modules is set to four, and the hidden dimension of all embeddings (excluding the initialization) has been standardized to 128. For the standard model, we choose the hyperparameter $\lambda$ in Query Adaptation as $\lambda=0.1$ for NELL and $\lambda=0.5$ for FB15K-237 and ConceptNet.  All methods have utilized 100-dimensional relation and entity embeddings.

For pretrained embeddings, we have employed TransE~\cite{transe} for the NELL and FB15K-237 datasets, while ComplEx~\cite{complex} has been utilized for ConceptNet. In the context of the NELL dataset, the TransE embeddings have been integrated by concatenating $v_{head}-v_{tail}$ to $E_s$ and $E_q$ within the \emph{Query Adaptation} phase. Here, $v_{head}$ and $v_{tail}$ signify the pretrained embeddings of the head and tail entities, and an optional neural network ($NN(v_{head}-v_{tail})$) can also be added. For the FB15K-237 dataset, a $BatchNorm$ Layer has been introduced within the $Linear$ layer in Eq.~(\ref{equation7}).

Regarding optimization, we have employed AdamW~\cite{adamw} with the learning rate $10^{-5}$, utilizing a linear schedule with 2,000 warm-up steps and a total of 20,000 steps.

To ensure robustness and reliability, each reported experimental result is derived from the average value obtained through conducting three independent experiments.

\subsection{Experimental Details}
We conduct all our SAFER training and testing procedures using NVIDIA RTX A6000 GPUs with a memory capacity of 48GB. Each training and testing instance was executed on a single GPU, and conducted using Python 3.10.10. We implement our framework with PyTorch.

\subsection{Limitations}
In this section, we introduce the limitations of our work in detail. Our SAFER model incorporates the Query Adaptation (QA) module to mitigate the inclusion of spurious information derived from the Support Adaptation (SA) module. For tail candidates with notably high scores, indicating substantial similarity between query and support graphs, the presence of extracted spurious information can severely impact the scoring process. In this way, the model tends to compare the most important and detailed information between support and query. Consequently, this has resulted in a remarkable enhancement in Mean Reciprocal Rank (MRR) and Hits@1 metrics.

However, this adaptation process inadvertently can still lead to the omission of certain global information from the support graph. This is a consequence of transferring all support information for processing onto the query graph. Consequently, the improvements of SAFER in Hits@5 and Hits@10 metrics are not as pronounced as those observed in MRR and Hits@1.

At present, we have yet to devise a solution to effectively integrate global information into predictions. Balancing the incorporation of detailed and global information concurrently presents a challenge that necessitates further investigation and future research endeavors.

\end{document}